\pdfoutput=1

\documentclass[11pt]{article}

\usepackage{acl}

\usepackage{times}
\usepackage{latexsym}

\usepackage{bm}
\usepackage{amsfonts}
\usepackage{amsmath}
\usepackage{mathrsfs}
\usepackage{graphicx}
\usepackage{booktabs}
\usepackage{multirow}
\usepackage{float}
\usepackage{subcaption}
\usepackage{stfloats}
\usepackage{amssymb}
\usepackage{xparse}
\usepackage{booktabs}
\usepackage{colortbl}
\usepackage{xspace}
\usepackage{enumitem}
\usepackage{todonotes}
\usepackage{bbm}

\usepackage{lscape}

\usepackage{lscape}
\usepackage{longtable}
\usepackage{tcolorbox}
\newcounter{mybox}

\usepackage{soul} 
\usepackage{xcolor} 

\definecolor{lightblue}{RGB}{236,244,255}

\usepackage{array}

\usepackage[normalem]{ulem}
\useunder{\uline}{\ul}{}

\usepackage{pifont}

\usepackage[T5]{fontenc}

\usepackage[utf8]{inputenc}

\usepackage{microtype}

\usepackage{inconsolata}

\usepackage{graphicx}

\usepackage{mdframed}
\newmdenv[
  linewidth=0.5pt,
  roundcorner=2pt,
  backgroundcolor=gray!10,
  skipabove=5pt,
  skipbelow=5pt,
  leftmargin=0pt,
  rightmargin=0pt,
  innerleftmargin=5pt,
  innerrightmargin=5pt
]{notebox}

\usepackage{soul}

\definecolor{bgimmediacy}{RGB}{243, 233, 255}  
\definecolor{bgembodiment}{RGB}{230, 242, 242} 
\definecolor{bgagency}{RGB}{230, 240, 255}  
\definecolor{bgfaithfulness}{RGB}{230, 255, 230}  
\definecolor{bgclarity}{RGB}{255, 245, 230}  
\definecolor{bgeaseofcomfort}{RGB}{255, 230, 230}  

\newcommand{\keyphraseimmediacy}[1]{%
  {\sethlcolor{bgimmediacy}\hl{\textbf{#1}}}%
}
\newcommand{\keyphraseembodiment}[1]{%
  {\sethlcolor{bgembodiment}\hl{\textbf{#1}}}%
}
\newcommand{\keyphraseagency}[1]{%
  {\sethlcolor{bgagency}\hl{\textbf{#1}}}%
}
\newcommand{\keyphrasefaithfulness}[1]{%
  {\sethlcolor{bgfaithfulness}\hl{\textbf{#1}}}%
}
\newcommand{\keyphraseclarity}[1]{%
  {\sethlcolor{bgclarity}\hl{\textbf{#1}}}%
}
\newcommand{\keyphraseeaseofcomfort}[1]{%
  {\sethlcolor{bgeaseofcomfort}\hl{\textbf{#1}}}%
}

\title{Toward Machine Interpreting: \\ Lessons from Human Interpreting Studies}

\author{Matthias Sperber\quad Maureen de Seyssel\quad  Jiajun Bao\quad Matthias Paulik \\ 
Apple\\
\texttt{\{sperber, mdeseyssel, jbao3, mpaulik\}@apple.com}
}

\begin{document}
\maketitle


\begin{abstract}

Current speech translation systems, while having achieved impressive accuracies, are rather static in their behavior and do not adapt to real-world situations in ways human interpreters do. In order to improve their practical usefulness and enable interpreting-like experiences, a precise understanding of the nature of human interpreting is crucial. To this end, we discuss human interpreting literature from the perspective of the machine translation field, while considering both operational and qualitative aspects. We identify implications for the development of speech translation systems and argue that there is great potential to adopt many human interpreting principles using recent modeling techniques. We hope that our findings provide inspiration for closing the perceived usability gap, and can motivate progress toward true machine interpreting.

\end{abstract}

\section{Introduction}

Even though speech translation~(ST) research has celebrated great successes, the user experience when employing ST technology in real-world tasks is often still perceived to be inferior to the experience of receiving assistance from a human interpreter \cite{iwslt2024panel}. This subjective impression is in contrast to the impressive accuracies reported on standard benchmarks. For example, \newcite{wein2024barriers} report superiority to human interpreters as measured by common machine translation~(MT) metrics against reference translations, and \newcite{cheng2024towards} report parity when measured by their proposed valid information proportion metric. Such findings imply that the gap in user experience largely stems from factors not captured by such benchmarks \cite{savoldi2025translation}. Such factors likely include, among others, interpreters' flexibility in modes of operation, their situational awareness, advanced translation strategies that include cultural adaptation, effective error prevention or recovery strategies \cite{jones2002conference}. Many of these characteristics and well-studied features of human interpreting have obtained little attention from MT researchers, perhaps partly because technical solutions to emulate human interpretation used to be out of reach.

Recent advances on large language models~(LLMs) and their application to translation opens up an avenue towards closing the usability gap between ST\footnote{In this paper, we refer to MT as automatic translation between any modality (speech or text), and ST as automatic translation from speech to any modality.} and human interpretation. For example, LLMs with long context may allow accumulating ``increased knowledge about a communicative event'' \cite{fantinuoli2024situational} in its entirety, allowing systems to effectively mimic situational awareness. Multimodal LLMs may learn to leverage audiovisual context, thereby extending situational awareness beyond the spoken word \cite{yin2024survey}. Prompts can be engineered to flexibly inform the translation model about speaker/listener relationship and their cultural/topical knowledge gap, allowing for more helpful translations and appropriate cultural adaptation \cite{yao2024benchmarking}. Reinforcement learning and instruction tuning may enable systems to imitate human error prevention or recovery strategies \cite{goldberg2023reinforcement}.

To effectively make progress towards translation systems that provide a more pleasant, interpreting-like experience, a precise understanding of the nature and goals of human interpreting is crucial. To this end, this paper discusses human interpreting literature and draws out implications and opportunities for MT research.  We hope that our work will serve as inspiration on the quest of advancing current ST technology toward a more interpreting-like experience, i.e.~toward what we might call ``machine interpreting''.\footnote{We use this term cautiously: even though \textit{machine interpreting} has occasionally been used as a synonym of \textit{speech-to-speech translation}~(S2ST), it has been convincingly argued that given the current major differences between S2ST and human interpreting, true \textit{interpreting} is not something that machines currently achieve, hence the term \textit{machine interpreting} should be reserved for a future in which machines ``achieve credible performance in all aspects of embodied and situated cognitive processing'' \cite{horvath2021speech,pochhacker2024machine}.}

\section{Goals and Scope}
For our purposes, we understand \textit{interpreting} to mean real-time oral (speech-to-speech) translation. We focus on the well studied fields of simultaneous\footnote{Interpreting concurrently through a microphone/earphone setup.} interpreting~(SI) and consecutive\footnote{Taking turns with the speaker.} interpreting~(CI), leaving less formalized paradigms~(e.g.\ dialog interpreting) to the side, although many insights are more generally applicable.

We contend that the objective of machine interpreting should not be to uncritically replicate human interpreters, but rather to identify and emulate those aspects that are both desirable and feasible within the context of machine-based applications. For instance, interpreting research includes techniques meant to address purely human limitations affecting the interpreter, such as cognitive overload and exhaustion. The limitations machines face are different. We therefore will not focus on describing interpreting principles and techniques meant to address such human limitations.

We also note that while many of the discussed principles immediately raise questions regarding how these might be evaluated, a systematic treatment of evaluation is beyond this paper's scope.

\begin{table*}[t!]
    \centering
    \small
    \resizebox{\textwidth}{!}{
    \begin{tabular}{>{\raggedright\arraybackslash}p{80pt} p{180pt} p{160pt}}
        \toprule  
        \textbf{Feature} & \textbf{Description} & \textbf{Example} \\
        \midrule
        \cellcolor{bgimmediacy} Temporal immediacy
            & Produces interpretation in real-time.
            & Maintains 1–2 seconds ear-voice-span. \\
        \cellcolor{bgimmediacy} Spatial immediacy
            & Operates in proximity of speaker \& audience.
            & Shares stage with speaker.
            \\ 

        \cellcolor{bgembodiment} Multimodality
            & Uses visual or gesture cues when available.
            & Refers to chart while speaker points. \\

        \cellcolor{bgagency} Free/diverse actions
            & Dynamically adapts to any situation.
            & Adapts translation approach to content type. \\ 
        \cellcolor{bgagency} Interaction/influence
            & Acts as a independent agent when needed.
            & Requests clarification, improves acoustics. \\ 

        \cellcolor{bgfaithfulness} Intent translation
            & Interprets what is meant, not what is said.
            & Interpretation conveys hidden accusations. \\ 

        \cellcolor{bgfaithfulness} Interpreter uncertainty
            & Maintains trust by signaling own uncertainty.
            & “Speaker may have said ‘revenue’.” \\ 
        
        \cellcolor{bgfaithfulness} Speaker errors
            & Indicates or corrects unintentional speaker errors.
            & Corrects “million” to “billion” in context. \\ 
        
        \cellcolor{bgfaithfulness} Adaptation/explanation & Adapts or explains culture-specific expressions. & “Break a leg!” → “Good luck!”. \\ 
        

        \cellcolor{bgclarity} Explicitation & Explicitates logic, intent, order, viewpoints. & “..., according to X's view.” \\ 

        \cellcolor{bgclarity} Brevity & Keeps sentences short and clear. & “The results were strong. More tests needed.” \\

        \cellcolor{bgclarity} Rhetoric quality & Delivers exceptionally high rhetoric quality. & Adapts style to particular audience. \\ 
        


        \cellcolor{bgeaseofcomfort} Pleasant experience
            & Works reliably; pleasant voice; friendly eye contact.
            & Avoids hectic speech when falling behind. \\
        \cellcolor{bgeaseofcomfort} Cognitive ergonomics
            & Minimizes audience stress and fatigue.
            & Avoids complex language. \\

        \bottomrule
    \end{tabular}
    }
    \caption{Summary of operational (\sethlcolor{bgimmediacy}\hl{immediacy}, \sethlcolor{bgembodiment}\hl{embodiment}, \sethlcolor{bgagency}\hl{agency}) and qualitative (\sethlcolor{bgfaithfulness}\hl{faithfulness}, \sethlcolor{bgclarity}\hl{clarity}, \sethlcolor{bgeaseofcomfort}\hl{ease~of~comfort}) interpreting goals.} 
    \label{tab:summary-table}
\end{table*}

\subsection{A Note on Prior Interdisciplinary Work}
Although MT and human interpreting research have progressed largely independently \cite{pochhacker2024machine}, the possibility of learning from human interpreters has already been discussed in ST literature \cite[etc.]{paulik2010learning,shimizu2013constructing,grissomii2014dont,cheng2024towards,wein2024barriers}. In contrast to these prior works, which primarily focus on specific interpreting strategies, we wish to put additional emphasis on the deeper underlying principles that drive such interpreting strategies. We argue that only focusing on specifics is too limiting, for several reasons: (1)~some strategies (e.g., passivization) are highly language dependent and hard to generalize. (2)~A strategy-only view  tends to over-simplify the nature of human interpreting, also because (3)~it is difficult to exhaustively list all strategies employed by interpreters, as evidenced by the significant number of non-overlapping strategies mentioned in above papers.

\section{Features of Interpreting}\label{general-features-of-interpreting}

To categorize the characteristics of interpreting, and discuss their implications for MT, we will employ two orthogonal descriptive systems: first, a set of operational features proposed by \newcite{pochhacker2024machine} that sheds light on the ``how'' of interpreting; second, a set of complementary features following \newcite{jones2002conference} that characterize high quality of the rendered interpretation itself, the ``what'' of interpreting. Both descriptive systems (summarized in Table~\ref{tab:summary-table}, illustrated in detail in Appendix~\ref{app:human-interpreting-examples}) characterize the differences between human interpreting goals and existing ST solutions, contributing to our goal of identifying aspects that true machine interpreting systems would be expected to address. 
The development of such systems will require both the integration of suitable existing methods, and addressing unsolved problems. Accompanying this section, Table~\ref{tab:prior_work} therefore summarizes prior work and suggests areas for future research.

\subsection{Operational View}\label{operational-view}

\citet{pochhacker2024machine} propose that the defining features of interpreting (human or machine) are high degrees of \emph{immediacy}, \emph{embodiment}, and \emph{agency}. 
This definition purposefully goes further than most prior work and highlights that besides the often discussed immediacy aspect, interpreting is marked by additional core characteristics, the absence of which in ST systems partially explains the usability gap that exists despite MT having already achieved high degrees of immediacy (e.g.\ low latency).

\subsubsection{Immediacy}\label{immediacy}

Interpreting is characterized by a high degree of immediacy in both a temporal and a spatial sense.  \emph{Temporal} immediacy requires that the pace of translation be determined by the source speaker and the interaction between speaker and recipient, not by the translator \cite{pochhacker2024machine}. Temporal immediacy is usually referred to as \emph{real-time} translation in MT literature \cite{papi2024real}. \emph{Spatial} immediacy indicates that the interpreter should be physically present at the location of the communicative event \cite{jones2002conference}. Immediacy is both a desirable property and a limiting factor: real-time cross-lingual communication in a particular place at a particular time are of high utility, while also limiting the interpreter's scope for repair or revision \cite{kade1968zufall}.  
Note that immediacy is critical in both SI and CI. SI tends to emphasize the temporal aspect, requiring results within a few seconds, and CI tends to emphasize the spatial aspect, with the interpreter often standing right next to the speaker.
In MT, spatial immediacy might easily be achieved through a portable device \cite{eck2010jibbigo}, while temporal immediacy requires efficient algorithms and hardware -- even in the less demanding consecutive setting.

\paragraph{Temporal immediacy in CI.}
In this scenario, the speaker and interpreter typically stand side by side and take turns, with the interpreter rendering the contents of the speaker's previous turn into the target language. Turns can be individual sentences or longer speech fragments of arbitrary length (15-minute fragments or longer are not uncommon). 
Turn taking between speaker and interpreter increases the duration of a speech considerably, hence interpreters are expected to deliver the interpreted speech as efficiently as possible. Specifically, interpreters \keyphraseimmediacy{speak immediately} when it is their turn, and aim at delivering a speech that is \keyphraseimmediacy{shorter and more concise} than the original speech \cite{poechhacker2012consecutive}. A good rule of thumb is to aim for 75\% of the source speech duration, although the ideal ratio depends on many factors, such as the speed and verbosity of the source speech \cite{jones2002conference}. 
Consecutive machine interpreting is relevant in situations where simultaneous interpretation via parallel channels is not feasible. 

\paragraph{Temporal immediacy in SI.}
Here, interpreters speak concurrently with the source speaker, typically equipped with a soundproofing booth for the interpreter, a microphone, and earphones for the audience. Interpreters navigate an \keyphraseimmediacy{ideal voice-ear-span} between too low and too high \cite{janikowski2025evs}. Interpreting at extremely low latency deteriorates quality by pushing interpreters toward unnaturally sounding translationese and toward making errors due to the inadequate context. But interpreting at overly high latency also deteriorates quality: it increases the risk for forgetting contents of the speech, and may result in accumulated latencies over time, often followed by compromised quality as interpreters rush to catch up. 

To help interpreters balance this trade-off, \newcite{jones2002conference} recommends latencies substantially lower than 5 seconds,\footnote{Empirically, average interpreter ear-voice-spans ranging between 2 and 5 seconds \cite{seeber2011cognitive} have been reported.} and outlines several principles: interpreters should (1) speak as soon as possible, (2) aim at grammatical speech with natural pauses not mid sentence but between completed sentences,\footnote{Relatedly, \newcite{macias2006probing} cautions that lengthy pauses (>2 seconds) may unintentionally be perceived as disfluent speech or omission errors by listeners.} and (3) start speaking only when a semantic unit is completely available.  
These principles can be summarized pragmatically by stating that interpreters should \keyphraseimmediacy{wait to speak until confident that the sentence can be finished without making unnatural breaks} caused by waiting for additional information, a principle that has been partly modeled in MT systems trained on segmented data from human interpreters~\cite{nakabayashi2019simulating} (this does not mean that all information must be available when \textit{starting} to speak the interpreted sentence). Speaking in short sentences is an effective way to reduce latency in this context. We note that the summarization principle may be especially applicable for machine interpreting purposes because it is straightforward to operationalize, such that it could aid the design of simultaneous speech-to-speech translation systems, or of streaming text-to-speech modules that are used in a cascade on top of simultaneously generated text output. 

There are two situations in which the interpreter aims for \keyphraseimmediacy{especially low latency: the beginning of the speech, and the end of the speech} \cite{jones2002conference}. At the beginning of the speech, starting to interpret immediately is important because it signals listeners that the interpreter is ready to operate, and listeners do not need to worry about missing any contents. In this case, it is even permissible for interpreters to invent light phrases (``hello'', ``ladies and gentlemen'', etc.) that the speaker did not actually say, just in order to say \textit{something}. At the end of the speech, low latency is important because there may be actions to take immediately after the speech (applause, preparing replies or questions, moving around the room, etc.), such that waiting for several seconds until the interpreted speech ends may put listeners in an awkward position.

In order to achieve low latency while maintaining high quality, interpreters aim to choose \keyphraseimmediacy{simple sentence structures} that provide flexibility and control in how one might finish the sentence \cite{jones2002conference}. For instance, interpreters avoid starting sentences with relative or subordinate clauses, because this limits options for continuing the sentence. Other strategies exist to minimize latency which may come at the cost of compromising quality or control and must therefore be used with care. Examples are passivization, generalization (replace ``spin dryer, cooker, and vacuum cleaner'' by ``electrical appliances''), omission of non-crucial contents, and anticipation of future content. Such compromises can be preferable over situations in which interpreters end up producing overly fast or hectic speech \cite{chmiel2024syntax}. 

\subsubsection{Embodiment}\label{embodiment}

\citet{pochhacker2024machine} introduce embodiment as referring to factors including \keyphraseembodiment{spatial/geographic situatedness}, and usage of \keyphraseembodiment{multimodal} communication channels. 
Human interpreters naturally base translations not just on the speaker's words, but utilize the full multimodal context. This includes reading the speaker's body language, visual cues, shared context regarding the nature of the communicative event and of the geographical location, or of any writing and diagrams that are available on slides or elsewhere in physical space, all of which might explain or disambiguate the communicative intent \cite{arbona2024role}. Moreover, interpreters will themselves actively use body language and facial expressions to clarify their interpretation, to signal readiness or technical problems, or to navigate speech discourse in the case of CI \cite{ahrens2004nonverbal}.

It is clear that human-level embodiment, including aspects such as spatial situatedness, input multimodality, and output multimodality, is tremendously difficult for machines to achieve. Although initial steps have been made toward supporting multimodal inputs \cite{caglayan2020simultaneous}, progress is hindered by lack of data and sparsity of visual and other sensory signal in practice. Moreover, we may also wish to render \textit{outputs} in an embodied or multimodal fashion. Machine interpreting could be seen both at a disadvantage and at an advantage in this regard: on the one hand, unless one employs a humanoid robot or avatar \cite{xie2015expressive}, the MT system lacks a body and can therefore not employ gestures and facial expressions. On the other hand, it can display information on screen concurrently to generating speech, in order to convey additional (or redundant) information, signal readiness, or call out technical problems. It may employ non-speech sounds for the same purpose, open multiple speech channels in parallel  
through headphones, or signal which of several source speakers is currently being interpreted by cloning the speaker's voice, to name just a few design options.

\subsubsection{Agency}\label{agency}

Human interpreters possess a high degree of agency \cite{llewellyn-jones2013getting}: they may choose to work in consecutive or simultaneous fashion, and may spontaneously switch, e.g., in case of technical problems. They might switch from merely bridging the language barrier to addressing knowledge gaps between speaker and recipient, e.g., by adding short explanations.
They may choose to ask questions of clarification back to the speaker, and can even refuse to interpret in case of inadequate acoustic conditions. They are also required to exercise good judgement in case the speaker makes errors: if the error is unintended (e.g., a number or term that's clearly wrong and unintended given the context), a good interpreter would silently correct the error, but only insofar as this provides no embarrassment for the speaker. Interpreting cannot easily be reduced to a fixed set of behaviors and techniques, because the number of unique situations to which the interpreter would act spontaneously and fittingly is unbounded.

\citet{pochhacker2024machine} elaborates that ``agency strongly implies intentionality and would therefore be closely associated with humanness, but it can accommodate the view that a degree of agency can also be attributed to machines''. He refers to \newcite{engen2016machine}, who 
generalize the concept of agency in a way applicable to both humans and machines, defining it as the capacity to perform activities in a particular environment according to certain objectives. The degree of agency would then strongly be determined by (a) the actions that can be performed, (b) their type (to what degree are these \keyphraseagency{actions free and diverse}?), and (c) the \keyphraseagency{ability to interact with and influence} other actors.

Current ST systems are highly restricted in their available actions and ability to influence. Existing efforts include multilingual models (actions are language pairs), automatic voice activity detection (actions are deciding when to start/stop translating). Controllable MT also increases the number of available actions, but usually needs to be triggered by users manually. However, such manually devised modeling approaches are unlikely to scale to the degree of agency expected by interpreters, and more flexible and scalable approaches are needed.

\subsection{Qualitative View}\label{qualitative-view}

Quality in MT is traditionally understood as achieving similarity to a human-created reference translation, a perspective which might lead researchers to ignore some quality aspects that are highly relevant to interpreting. Per \citet{jones2002conference}, the interpreter's overarching goal can be summarized as being threefold, namely interpreting ``(1) with greatest faithfulness to the original but also (2) greatest clarity and (3) ease of comfort for the listener''. 
Here we understand faithfulness as related to meaning and speaker intent (both broad and subtle), clarity as related to wording and voicing for optimal comprehensibility, and ease of comfort as related to form of presentation. While interpreters will strive simultaneously for optimal faithfulness, clarity, and ease of comfort, and while there is much overlap between the three goals, there may also at times be tension between these goals, requiring the interpreter to take reasonable trade-offs.\footnote{This is reminiscent of the trade-off between accuracy and fluency in traditional MT \cite{lim2024simpsons}.} At the same time, it is important to realize that interpreted speech can be of \emph{higher} quality than the source speech, e.g., by bringing additional clarity that was lacking in the source speech. In the following, we will discuss each of the three aspects, heavily borrowing from \newcite{jones2002conference}'s view on the subject.

\subsubsection{Faithfulness}\label{sec:faithfulness}

Faithfulness is related to the familiar concept of accuracy (or adequacy) in MT literature \cite{white1993evaluation}, but takes this concept quite far: interpreters aim to faithfully convey the \keyphrasefaithfulness{speaker's intent}, i.e., what source speaker \textit{meant} matters more than what they said.\footnote{Faithfulness as understood here is quite different from the faithful translation approach of \newcite{newmark1981approaches}, which falls closer to the literal side of the translation spectrum.} This means that at the heart of faithful interpreting lies the apparent paradox that ``in order to be faithful to the speaker, the interpreter must betray them'' \cite{jones2002conference}. 

One common way in which a faithful interpreter is expected to deviate from what the speaker said (but not what they meant) regards cultural references. The audience may not be familiar with certain places, public figures, currencies, conversion units, commonly used metaphors, etc. There is then a choice between literal translation, resorting to a non-literal equivalent in the target language \keyphrasefaithfulness{(adaptation), and/or explanation}. In many cases, explanations are preferable \cite{jones2002conference}, but the level of detail can be tricky to get right: the interpreter must include just enough detail to convey the speaker's intent understandably to the particular audience, without adding so much explanation as to distract from the speaker's point. 

Another common example for when the interpreter is expected to deviate from what the speaker said is when the speaker \keyphrasefaithfulness{unintentionally misspeaks} \cite{besien2014dealing}. \citet{jones2002conference} recommends interpreters not to simply translate such mistakes as-is, because the audience may think that the error was made by the interpreter instead of the speaker, causing the audience to lose trust in the interpreter. Instead, the interpreter must choose to either silently correct the error, or to explicitly state uncertainty (but in a manner that does not embarrass the source speaker). Moreover, in some cases interpreters may tone down rude remarks which the speaker regrets as soon as having uttered them, which are in that sense unintentional. On the other hand, any deliberate speech (including flawed logical arguments, impoliteness, dishonesty) must always be translated unchanged.

As a general rule, faithful interpretation means that the interpreter should say what the source speaker \emph{would have said} in the target language, if (s)he were fluent in that language. This requires interpreters to overcome \keyphrasefaithfulness{specific linguistic challenges} that are too numerous to list here. Examples of such challenges include the need to compile a technical glossary ahead of time in preparation of technical content interpretation, paying special attention to nuances in the source speech, staying consistent in style and vocabulary in the context of long speeches, etc.\\

It must be noted that a critical requirement in faithful interpretation is that the interpreter is trusted by the listener (and speaker) to always be rendering a reliable interpretation and that the interpreter resists the temptation of coloring their interpreted speech with any agenda, opinions and world view of their own. The interpreter establishes this trust both by adhering to a high standard of accuracy, and by transparently admitting \keyphrasefaithfulness{uncertainty/error}. Whenever in doubt, instead of guessing, interpreters adhere to a sort of error handling cascade: first, aim for perfection; if uncertain about minor details, simplify or generalize; if uncertain about important content, ask clarification questions to the source speaker if possible (usually in consecutive mode), or else admit failure; if failure becomes too frequent due to poor interpreting conditions (e.g.~acoustic), warn the audience about it or even refuse to interpret until conditions are improved \cite{jones2002conference}.\\

Conceivably, many of the particular interpreting strategies discussed (establishing trust through error handling cascades, toning down rude comments, adding explanations, etc.) could be solved through available techniques such as prompting tuning or preference optimization \cite{yu2025simulpl}. An open question is whether these rules are too numerous to solve through individual strategies, in which case one may resort either to data driven approaches or to designing simple overarching prompts that implement general principles (e.g.~rendering what the source speaker would have said in the target language). However, the biggest challenge may lie in choosing \emph{when} to apply certain solutions and when not to. This brings us back to the notion of agency discussed in the previous section: machine interpreting systems would be required to proactively select appropriate translation and error recovery strategies depending on circumstances, a desideratum that is currently only achieved by human interpreters (and tends to be the main factor that distinguishes an excellent interpreter from a good interpreter). Even assuming that a system could be designed that successfully mimics a skillful human interpreter in all of these aspects, the question of whether or not this is even desirable must be addressed: perhaps most users would want silent correction of unintentional minor mistakes only from a human interpreter but not a machine? Or perhaps only if indicated on screen and not in fact silent?

\subsubsection{Clarity}\label{sec:clarity}
\begin{table*}[tb!]
  \centering
  \scriptsize
  \begin{tabular}{p{2.6cm}p{4.0cm}p{4.0cm}p{3.7cm}}
    \toprule
    \textit{Principle} & \textit{Existing MT research} & \textit{Research in related fields} & \textit{Potential research gaps} \\
      \midrule
        
      \multicolumn{4}{c}{\textbf{Temporal immediacy (consecutive case)}}  \\
        
        \keyphraseimmediacy{Speak immediately} 
            & Efficient decoding \cite{junczys-dowmunt2018marian}, model compression \cite{rajkhowa2025semi}.
            & Low delay endpointing \cite{li2002robust,zink2024predictive}; recovery from false endpointing triggers \cite{ma2024language}.
            & General inference efficiency; Explore simultaneous MT techniques $\rightarrow$ speak before decoding finished. \\
 
        \keyphraseimmediacy{Short, concise speech}
            & Summarization \cite{bouamor2013sumt,karande2025crosslingual}; length control \cite{lakew2019controlling}. 
            & Speech-worthy language generation \cite{cho2024speechworthy}. 
            & Techniques exist but are not commonly applied in practice; user preferences? \\
      \hline
        
      \multicolumn{4}{c}{\textbf{Temporal immediacy (simultaneous case)}}  \\
        
        \keyphraseimmediacy{Ideal voice-ear-span}
            & Simultaneous S2TT \cite{fuegen2009thesis}, S2ST \cite{zheng2020fluent,ma2024nonautoregressive}; mimic interpreters \cite{grissomii2014dont,nakabayashi2019simulating}.
            & Incremental TTS \cite{liu2022start}, concurrent speech/text generation \cite{yang2024interleaved,fang2024llama-omni}.
            & Low latency S2ST methods less mature than S2TT, some room for improvement; lacking studies measuring user preference. \\

        \keyphraseimmediacy{Zero initial/final lag}
            & \multicolumn{1}{c}{--}
            & \multicolumn{1}{c}{--}
            & Likely unaddressed. \\
        
        \keyphraseimmediacy{No unnatural breaks}
            & Semantic segmentation \cite{huang2023speech}.
            & \multicolumn{1}{c}{--}
            & Related work exists, core issue likely unaddressed. \\
        \keyphraseimmediacy{Simple sentence structure}
            & Data driven approaches \cite{shimizu2013constructing,cheng2024towards}; preference learning \cite{yu2025simulpl}.
            & \multicolumn{1}{c}{--}
            & Initial work exists. \\
      \hline
        
      \multicolumn{4}{c}{\textbf{Embodiment}}  \\
        
        \keyphraseembodiment{Multimodality}
            & Multimodal MT \cite{caglayan2020simultaneous}, directional audio \cite{chen2025spatial}.
            & Speech-to-pictogram \cite{macaire2024towards}; avatar animation \cite{xie2015expressive}.
            & Large open-ended research space. \\
      \hline
         
      \multicolumn{4}{c}{\textbf{Agency}}  \\
        \keyphraseagency{Free/diverse actions, \\interaction}
            & Voice activity detection \cite{graf2015features}; controllable MT \cite{agrawal2019controlling}.
            & Interactive speech LLMs \cite{li2025baichuan}.
            & Promising techniques exist but need exploration and integration with MT. \\
      \hline
        
      \multicolumn{4}{c}{\textbf{Faithfulness}}  \\
        \keyphrasefaithfulness{Intent translation} 
            & Non-linear translation with LLMs \cite{yao2024benchmarking}; prosodic intent \cite{anumanchipalli2012intent}, lexical choice \cite{tsiamas2024speech}; expressive S2ST \cite{gong2024seamlessexpressivelm}. 
            & Intent discovery \cite{song2023large}.
            & Promising techniques exist, progress potentially hindered by established evaluation methods not rewarding shift toward non-linear translation. \\
        \keyphrasefaithfulness{Interpreter uncertainty} 
            & Quality estimation: segment level \cite{specia2010combining}, speech \cite{han2024speechqe}; trust \cite{savoldi2025translation}; uncertainty disentangling \cite{zerva2022disentangling}.
            & Confidence estimation \cite{papadopoulos2001confidence}; generating ``answer unknown'' through reinforcement learning \cite{goldberg2023reinforcement}.
            & Plenty of related research, but ST angle underexplored. \\
        \keyphrasefaithfulness{Speaker errors} 
            & Robust MT \cite{anastasopoulos2019neural}; error correction \cite{koneru2024blending}.
            & Factual text correction \cite{shah2020automatic}.
            & Sensitive topic with initial work, needs careful investigation. \\
        \keyphrasefaithfulness{Adaptation/explanation} 
            & Cultural adaptation \cite{yong2024cultural,conia2024towards}; named entity explanation \cite{han2023bridging,peskov2021adapting}.
            & Culturally aligned LLMs \cite{li2024culturellm}.
            & More holistic methods and user studies needed. \\
      \hline
        
      \multicolumn{4}{c}{\textbf{Clarity}}  \\
        \keyphraseclarity{Discourse/intent explicitation} 
            & Labeled data \cite{meyer2013machine}; corpus analysis \cite{lapshinova2022exploring}. 
            & (Chrono)logic argument \cite{hulpus2019towards,mendoza2024translating}, humor \cite{peyrard2021laughing} detection.
            & Not yet addressed directly, limited related work exists. \\
        \keyphraseclarity{Short, easy-to-digest \\sentences} 
            & Simplification for text \cite{oshika2024simplifying}, speech \cite{wu2025improve} translation; disfluency removal \cite{cho2016multilingual,salesky2019fluent}.
            & Text \cite{laban2021keep}, spoken language \cite{cho2024speechworthy} simplification.
            & Initial work exists, need more investigation, user studies, application in practice. \\
        \keyphraseclarity{Rhetoric quality} 
            & Expressive S2ST \cite{huang2023holistic}; on-the-fly adaptation. \cite{morishita2022domain}. 
            & Expressive TTS \cite{cohn2021prosodic}. 
            & Recent work exists, but needs improvement especially in simultaneous scenario. \\
      \hline
        
      \multicolumn{4}{c}{\textbf{Ease of Comfort}}  \\
        \keyphraseeaseofcomfort{Pleasant listening experience} 
            & User-centric MT \cite{liebling2020unmet,briva-iglesias2023impact}; eval.\ through questionnaires \cite{mueller2016evaluation}. 
            & HCI design principles \cite{norman1983design}; cross-lingual voice cloning \cite{zhang2019learning}.
            & Underexplored. \\
        \keyphraseeaseofcomfort{Cognitive ergonomics} 
            & Translation reading assistance \cite{minas2025adaptive}; evaluation through interviews \cite{mueller2016evaluation}, eye tracking \cite{castilho2018reading,guerberof2021impact}.
            & Reduce cognitive strain through multimodality \cite{malakul2023effects}; LLM cognitive ergonomics \cite{wasi2024cogergllm}. 
            & Underexplored. \\
    \bottomrule
  \end{tabular}
  \caption{Exemplary prior work in MT and adjacent fields addressing selected characteristics. New abbreviations: S2TT (speech-to-text translation), TTS (text-to-speech synthesis).}
  \label{tab:prior_work}
\end{table*}

Given a source sentence $X$ and target sentence $Y$, it has been suggested to regard adequacy as corresponding to the conditional probability $p(X|Y)$ and fluency\footnote{Also referred to as intelligibility or well-formedness by \citet{white1993evaluation}.} as $p(Y)$ \cite{teich2020translation}. The same could be said of faithfulness and clarity, respectively, but the relationship is less direct: faithfulness goes beyond accuracy (see previous section), and similarly clarity also goes beyond fluency and requires an active effort of producing exceptionally clear speech. We might appropriately call such clear speech ``interpretese'', in a positive sense.

Clarity is achieved through clear pronunciation and well-formed language, but importantly also by \keyphraseclarity{explicitating intent and discourse} \cite{gumul2017explicitation}. Clear, well-formed and explicit speech counteracts inevitable comprehension gaps caused by the linguistic and cultural differences~\cite{meyer2013machine,meyer2013implicitation}. For instance, there is a risk that even skillfully interpreted jokes or persuasive rhetoric fail to carry their full intended effect, and clarity can be achieved by explicitly stating the speaker's intention to persuade, or to convey humor (``said the speaker jokingly''). Similarly, implicitly expressed points of view tend to get missed by listeners despite faithful interpretation, and need repeated clarification (``according to\ldots{}''). Implicit (chrono-)logical connections should also be made explicit (``first'', ``then''; ``therefore'', ``but'').

Clarity also demands that complex sentences be broken up into \keyphraseclarity{short, easy-to-digest sentences}. Short sentences help prevent interpreting  errors, but crucially also improve clarity and the overall listening experience. Speech becomes more comprehensible as complex sentence structures are replaced by simple grammar that conveys the speech intent more directly and accessibly. Short sentences can be obtained through splitting of long and complex sentences into shorter ones, but also through removing redundancy (e.g.\ rhetorical repetitions) and disfluent speech (e.g. hesitations).

Generalizing this idea further, interpreters aim at delivering an overall \keyphraseclarity{rhetorically good speech} \cite{jones2002conference}: using neither a bored nor overly intense but a natural intonation, avoiding disfluent speech and unnatural breaks, using effective placement of prosodic emphasis, speaking with a clear pronunciation and at a natural speed. Finally, clarity is improved through usage of terminology and expressions that the particular audience can relate with (e.g.~academic, political, casual).

On the MT side, many of the considerations discussed for faithfulness apply here as well, including the question of following data-driven, fine-grained strategies driven (most prior work), or overarching principles driven approaches; machine agency to invoke methods at appropriate moments; the need for appropriate evaluation and user studies. Wording/language generation and speech synthesis both contribute to clarity and most work hand in hand.

\subsubsection{Ease of Comfort}\label{sec:ease-of-comfort}

Good interpreters provide a \keyphraseeaseofcomfort{pleasant listening experience} for the audience \cite{kurz2001conference} and  \keyphraseeaseofcomfort{prevent cognitive strain}. While interpreting faithfully and with clarity contributes to these aspects, \textit{ease of comfort} includes additional nuances \cite{jones2002conference}. Among others, it is achieved through speaking with a pleasant voice, through establishing eye contact and a personal connection with the audience, and through signaling an ``I got you covered'' attitude. It also requires reliable and intuitive technical equipment such as headphones, volume control, and connectivity for the audience. Glitches in any of these areas would pose a distraction and cause a target-language audience to feel less well integrated and cared for than a source-language audience.

Ease of comfort is not a commonly used term in the MT community, and is more difficult to formalize than faithfulness and clarity. It is closer to a human-computer-interaction (HCI) research mindset and recognizes that a well-designed user interface (UI) is just as important to users as the quality of translations. While HCI work on MT is unfortunately sparse, human interpreting best practices may provide hints for what users' needs are, such as the engineering of robust and reliable ``I got you covered'' systems, and going out of one's way to reduce unnecessary cognitive strain from users which typically already find themselves struggling to navigate complex multicultural situations when employing speech translation tools.

\section{Discussion and Future Work}

Interpreting is a complex and multi-faceted activity, and the road to developing true machine interpreting systems that perform the nuanced communicative functions expected of human interpreters is difficult.
Among the many aspects discussed above, some are already successfully addressed (e.g.\ parts of immediacy); some are relatively low hanging fruit in the sense that while not usually integrated into MT systems, NLP methods exist to address them (e.g.\ summarization); some aspects are very challenging to address (e.g.\ agency, embodiment), but even for the challenging aspects, meaningful first steps are in sight thanks to recent progress with LLMs, multimodal representations, zero-shot learning, etc.\
As a starting point for identifying promising avenues for future work, Table~\ref{tab:prior_work} provides a (non-comprehensive) overview of research that partially addresses some of the identified characteristics, as well as potential research gaps.

It is important to note that not all discussed interpreting goals are equally relevant in all types of machine interpreting use cases: Some use cases may lend themselves to a more interactive design that benefits from sophisticated machine agency, but others may by nature be restricted in the degree of input/output multimodality, and some use cases may require taking stronger trade-offs such as prioritizing immediacy over other factors. User studies may facilitate such design choices. 

A common challenge with most of the discussed principles is that they are inherently difficult to evaluate, and may in fact detrimentally impact the established evaluation metrics based on similarity to reference translations. Even human evaluation will face difficulties uncovering all aspects, e.g.\ identifying an ideal immediacy-faithfulness tradeoff may require task-based evaluation for holistic assessment.
In addition, user studies that assessing whether certain interpreting principles are actually beneficial to users would be of high value. Such user studies need careful planning because users may not always be aware of their needs, as exemplified in a study on human interpreting that found users expressing no particular preference for natural intonation in the interpreted speech, but comprehension tests revealing a noticeable positive impact of good intonation \cite{shlesinger1994intonation}.

\section{Conclusion}\label{sec:conclusion}

This paper discussed human interpreting literature, with the aim of drawing implications for MT research and addressing what has been characterized as a ``lack of interaction and exchange" between the two research communities \cite{pochhacker2024machine}. Unlike prior ST work, we have focused on higher level ideals and goals as identified by the interpreting research community, which allows sidestepping the otherwise difficult question of which specific interpreter strategies should or should not be imitated by MT. We categorized insights into operational and qualitative axes, finding that among the various aspects, only immediacy can meet the standards of interpreting. At the same time, recent modeling advances such as general-purpose LLMs and multimodal learning seem to have paved the way toward making significant progress on many of the remaining fronts, placing the development of more user-centric ST systems within sight.

\newpage

\section*{Limitations}\label{sec:limitations}

This paper aimed to give a high level perspective of human and machine interpreting. While we hope to have covered all important aspects, by nature this vast topic cannot be comprehensively treated within the given space constraints. Among others, we have not discussed that some aspects are subject to debate among human interpreting researchers, such as the question on how eager interpreters should be to correct speaker errors. We have also centered discussions on \textit{conference} interpreting literature, mainly owing to the abundance of literature on this setting. While the operational and qualitative aspects generalize to most interpreting settings, covering literature on additional paradigms such as dialog interpreting may yield additional insights with regards to how to trade off and prioritize the discussed dimensions. On the technical side, the provided references on prior work are only a selective sample. For a systematic and comprehensive treatment of the technical aspects, we refer to overview papers such as \newcite{seligman2019advances,sperber2020speech,sulubacak2020multimodal,savoldi2025translation,sperber2019thesis}.

\bibliography{acl_latex}

\newpage

\appendix
\section{Human interpreting examples}\label{app:human-interpreting-examples}

This appendix provides a non-exhaustive list of concrete examples (Table \ref{tab:examples_interpretation_scenarios}) illustrating the principles of human interpreting discussed in the main paper. The examples demonstrate how specific interpreting phenomena relate to both the operational dimensions (immediacy, embodiment, agency) proposed by \citet{pochhacker2024machine} and the qualitative dimensions (faithfulness, clarity, ease of comfort) described by \citet{jones2002conference}.

This appendix grounds the concepts presented in the main paper  in practical scenarios that interpreters encounter. Each example shows how interpreters might have to make decisions that concurrently address multiple dimensions: for instance, how accounting for the type of audience in order to correctly adapt the text might encompass both embodiment and faithfulness. 

The table is organized into functional categories (information management, handling uncertainty/errors, cultural \& pragmatic adaptation, delivery \& fluency, and simultaneous MT specifics) to help comparison across different types of interpreting challenges.

\onecolumn



\renewcommand{\arraystretch}{2.2}
\tiny

\begin{longtable}{@{}
>{\raggedright\arraybackslash}p{1.5cm}
>{\raggedright\arraybackslash}p{2.2cm}
>{\raggedright\arraybackslash}p{0.8cm}
>{\raggedright\arraybackslash}p{0.8cm}
>{\raggedright\arraybackslash}p{0.6cm}
>{\raggedright\arraybackslash}p{4.4cm}
>{\raggedright\arraybackslash}p{4.4cm}
@{}}

\caption{Illustrative examples of linguistic and pragmatic phenomena observed in human interpreting, categorized according to both the operational and qualitative views. The \textit{mode} column indicates whether the phenomenon is primarily relevant to simultaneous interpreting (SI), consecutive interpreting (CI), or both modes. Examples show source text/speech and corresponding target interpretations.} \\

\toprule
\textbf{Phenomenon} & \textbf{Explanation} & \textbf{Op. view} & \textbf{Qual. view} & \textbf{Mode} & \textbf{Example (Source)} & \textbf{Example (Target)} \\
\midrule
\endfirsthead

\multicolumn{7}{@{}l}{\small\textit{(Continued from previous page)}} \\
\toprule
\textbf{Phenomenon} & \textbf{Description} & \textbf{P\"{o}chhacker Concepts} & \textbf{Jones Concepts} & \textbf{Mode} & \textbf{Example (Source)} & \textbf{Example (Target)} \\
\midrule
\endhead

\bottomrule
\multicolumn{7}{r}{\small\textit{(Continued on next page)}} \\
\endfoot

\bottomrule
\endlastfoot

\rowcolor[HTML]{F0F0F0}\multicolumn{7}{@{}l}{\textbf{Information Management}} \\[0.5em]

Compression (Simultaneous) &
Reducing the source content to its essential meaning while omitting detail. &
\sethlcolor{bgagency}\hl{Agency} & \sethlcolor{bgeaseofcomfort}\hl{Ease of comfort} & SI &
\parbox[t]{\linewidth}{
 \texttt{“The rapid increase in global temperatures, coupled with the alarming rise in sea levels, has created an unprecedented challenge for coastal communities.”}
} &
\parbox[t]{\linewidth}{
\texttt{“Rising global temps and sea levels challenge coasts.”}
} \\

Simplification / Restructuring &
Reformulating complex syntax or discourse into simpler, more accessible structures. &
\sethlcolor{bgagency}\hl{Agency}  & \sethlcolor{bgeaseofcomfort}\hl{Ease of comfort} & Both &
\parbox[t]{\linewidth}{
\texttt{“The man, who was wearing a hat, which was red, and had a feather, which was very long, walked into the room.”}
} &
\parbox[t]{\linewidth}{
\texttt{“The man wearing a red hat with a long feather walked into the room.”}
} \\

Making Implicit Content Explicit &
Making implied meaning explicit to ensure clarity for the listener.	 &
\sethlcolor{bgagency}\hl{Agency} & \sethlcolor{bgclarity}\hl{Clarity} & Both &
\parbox[t]{\linewidth}{
\textsc{\textbf{Spk 1}}:  \texttt{“Are you coming to the party?”} 
\newline\textsc{\textbf{Spk 2}}: \texttt{“I have to work.”}
}
 &
\parbox[t]{\linewidth}{
\textsc{\textbf{Spk 1}}:  \texttt{“Are you coming to the party?” }
\newline\textsc{\textbf{Spk 2}}:\texttt{ “No, I have to work.”}
} \\

Short, easy-to-digest phrases &
Segmenting output into brief, manageable chunks for ease of processing. &
\sethlcolor{bgagency}\hl{Agency} & \sethlcolor{bgclarity}\hl{Clarity} & Both &
\parbox[t]{\linewidth}{
\texttt{“The new climate report, which has been reviewed by over 200 scientists worldwide and covers a 30-year span, warns of unprecedented environmental changes.”}
} &
\parbox[t]{\linewidth}{
 \texttt{“The new climate report warns of major environmental change. It was reviewed by 200 scientists. It spans 30 years.”}
} \\

\rowcolor[HTML]{F0F0F0}\multicolumn{7}{@{}l}{\textbf{Handling Uncertainty / Errors}} \\[0.5em]

Speaker Uncertainty (in Source) &
Recognising and conveying the speaker’s lack of certainty. &
\sethlcolor{bgagency}\hl{Agency}  & \sethlcolor{bgfaithfulness}\hl{Faithfulness} & Both &
\parbox[t]{\linewidth}{
 \texttt{“The company reported... uh... I think it was a 10 or 12 percent... mumble... increase.”}
} &
\parbox[t]{\linewidth}{
\texttt{“The company reported a profit increase of approximately 10 to 12 percent.”}
} \\

Interpreter/MT Uncertainty &
Indicating hesitation, uncertainty about the input in the produced output. &
\sethlcolor{bgagency}\hl{Agency}  & \sethlcolor{bgfaithfulness}\hl{Faithfulness} & Both &
\parbox[t]{\linewidth}{
\textit{[Poor audio/mumble]}\hspace{1em}\texttt{“[...] reported} \textit{[...]} \texttt{percent increase...”}
} &
\parbox[t]{\linewidth}{
\texttt{“[...] reported some type of percent increase that I am unsure of...”}\hspace{1em}\textit{[Signals uncertainty]}
} \\

Factual Error (in Source) &
Correcting factual inaccuracies in the source speech during interpretation.	 &
\sethlcolor{bgagency}\hl{Agency} & \sethlcolor{bgfaithfulness}\hl{Faithfulness} & Both &
\parbox[t]{\linewidth}{
\texttt{“We’re excited to be in Paris. Aachen is such a nice city.”}
} &
\parbox[t]{\linewidth}{
 \texttt{“We’re excited to be in Paris. \textbf{Paris} is such a nice city.”}\hspace{1em}\textit{[Error corrected]}
} \\

Source Misspeaks / Self-Correction &
Conveying the speaker’s final, corrected intent while omitting false starts. &
\sethlcolor{bgagency}\hl{Agency}  & \sethlcolor{bgfaithfulness}\hl{Faithfulness} & Both &
\parbox[t]{\linewidth}{
\texttt{“The meeting is on Wednes- uh, Thursday”}
} &
\parbox[t]{\linewidth}{
\texttt{“The meeting is on Thursday”}
} \\

Resolving errors interactively	 &
Revising or correcting prior output in response to listener feedback or clarification.	 &
\sethlcolor{bgagency}\hl{Agency}  & \sethlcolor{bgfaithfulness}\hl{Faithfulness} & CI &
\parbox[t]{\linewidth}{
\textsc{\textbf{User}}: \hspace{1em}\texttt{“Turn left.”} \\ \textsc{\textbf{MT output}}: \hspace{1em}\texttt{“Turn right.”} \\ \textsc{\textbf{User}}:: \hspace{1em}\texttt{“No, left!”}
} &
\parbox[t]{\linewidth}{
\textsc{\textbf{MT output}}: \textit{[After previous utterances]}\hspace{1em}\texttt{“Sorry, turn left.”}}
 \\

\rowcolor[HTML]{F0F0F0}\multicolumn{7}{@{}l}{\textbf{Cultural \& Pragmatic Adaptation}} \\[0.5em]

Cultural References / Idioms &
Identifying culture-specific items/phrases and adapting them for target audience understanding. &
\sethlcolor{bgagency}\hl{Agency}  & \sethlcolor{bgfaithfulness}\hl{Faithfulness} & Both &
\parbox[t]{\linewidth}{
 \texttt{“He really moved the goalposts.”}
} &
\parbox[t]{\linewidth}{
\textbf{\textsc{[Option A]}}\hspace{1em}\texttt{“He changed the requirements unfairly.”}\\[0.3em]
\textbf{\textsc{[Option B]}}\hspace{1em}\texttt{“He moved the goalposts. In other words, he changed the rules unexpectedly.”}
} \\

Play on Words / Humor &
Reproducing the humour or rhetorical effect of a pun or wordplay using target-language strategies. &
\sethlcolor{bgagency}\hl{Agency}  & \sethlcolor{bgfaithfulness}\hl{Faithfulness} & Both &
\parbox[t]{\linewidth}{
\texttt{I'm reading a book about anti-gravity. It's impossible to put down!”}
} &
\parbox[t]{\linewidth}{
\textbf{\textsc{[Option A]}}\hspace{1em} \texttt{“I'm reading a book about anti-gravity. It's impossible to put down!”} \textit{[Verbatim if pun works in the target language]}\\[0.3em]
\textbf{\textsc{[Option B]}}\hspace{1em} \texttt{“I'm reading a book about time travel. It's about time!”} \textit{[Different pun suited to the target language]}
} \\

Understanding of Audience & Adjusting language, complexity, and detail based on the target audience. & \sethlcolor{bgembodiment}\hl{Embodiment} & \sethlcolor{bgfaithfulness}\hl{Faithfulness} & Both &
\parbox[t]{\linewidth}{ \texttt{“Gene expression was upregulated.”}} &
\parbox[t]{\linewidth}{\texttt{“The activity of the gene was increased.”}} \\

Handling Offensive/Sensitive Language &
Deciding whether/how to translate sensitive language based on context, audience, and ethics. &
\sethlcolor{bgagency}\hl{Agency}  & \sethlcolor{bgfaithfulness}\hl{Faithfulness} & Both &
\parbox[t]{\linewidth}{
 \texttt{“That was a *** disaster.”}
} &
\parbox[t]{\linewidth}{
\textbf{\textsc{[Option A]}}\hspace{1em} \texttt{“That was a total mess.”} \textit{[Mitigated]}\\[0.3em]
\textbf{\textsc{[Option B]}}\hspace{1em} \texttt{“That was a *** disaster.”} \textit{[Verbatim]}
} \\

Meaning conveyed through tone &
Identifying and conveying the source's rhetorical intent, tone, or style. &
\sethlcolor{bgagency}\hl{Agency} & \sethlcolor{bgfaithfulness}\hl{Faithfulness} & Both &
\parbox[t]{\linewidth}{
 \texttt{“Oh, that was a brilliant idea!”} \textit{[said with heavy sarcasm]}
} &
\parbox[t]{\linewidth}{
\textbf{\textsc{[Option A]}}\hspace{1em} \texttt{“Oh, that was a brilliant idea!”} \textit{[Maintains sarcastic tone]}\\[0.3em]
\textbf{\textsc{[Option B]}}\hspace{1em} \texttt{“Well, that was obviously a terrible idea.”} \textit{[Explicates the sarcasm]}
} \\

\rowcolor[HTML]{F0F0F0}\multicolumn{7}{@{}l}{\textbf{Delivery \& Fluency}} \\[0.5em]

Disfluent Speech / Handling Disfluencies & Smoothing, simplifying, or retaining speaker disfluencies depending on context. & \sethlcolor{bgagency}\hl{Agency} &  \sethlcolor{bgfaithfulness}\hl{Faithfulness} & Both &
\parbox[t]{\linewidth}{ \texttt{“I, um, was very ha-ha-happy to at- attend.”}} &
\parbox[t]{\linewidth}{ \texttt{“I was very happy to attend.”}} \\

Incorrect Linguistic Capabilities (Source) &
Choosing whether to correct or reproduce errors in the speaker’s grammar or usage. &
\sethlcolor{bgagency}\hl{Agency} & \sethlcolor{bgfaithfulness}\hl{Faithfulness} & Both &
\parbox[t]{\linewidth}{
 \texttt{The children \textbf{is} playing.”}
} &
\parbox[t]{\linewidth}{
\textbf{\textsc{[Option A]}}\hspace{1em} \texttt{“The children \textbf{are} playing.”} \textit{[Corrected]}\\[0.3em]
\textbf{\textsc{[Option B]}}\hspace{1em} \texttt{“The children \textbf{is} playing.”} \textit{[Retained to reflect speaker’s level]}
} \\

\rowcolor[HTML]{F0F0F0}\multicolumn{7}{@{}l}{\textbf{Simultaneous MT Specifics}} \\[0.5em]

Start speaking immediately &
Beginning interpretation output as soon as speech begins, without waiting for complete context. &
\sethlcolor{bgimmediacy}\hl{Immediacy} & \sethlcolor{bgeaseofcomfort}\hl{Ease of comfort} & SI &
\parbox[t]{\linewidth}{
 \textit{[The speaker begins talking without hesitation.]}
} &
\parbox[t]{\linewidth}{
\textit{[The interpreter or MT system begins output immediately, even before full sentence is heard.]}
} \\

Wait to speak until sentence can be finished &
Delaying output until enough source context is available to ensure accurate output. &
\sethlcolor{bgimmediacy}\hl{Immediacy} & \sethlcolor{bgclarity}\hl{Clarity} & SI &
\parbox[t]{\linewidth}{
 \textit{[Speaker begins with a fragment that could be misleading without the full sentence.]}
} &
\parbox[t]{\linewidth}{
 \textit{[Interpreter delays output until full sentence meaning is available.]}
} \\

Speak at low latency &
Maintaining a continuous, minimal-delay output that closely tracks the source speech. &
\sethlcolor{bgimmediacy}\hl{Immediacy} & \sethlcolor{bgeaseofcomfort}\hl{Ease of comfort} & SI &
\parbox[t]{\linewidth}{
\textit{[Speaker talks continuously with minimal pauses.]}
} &
\parbox[t]{\linewidth}{
 \textit{[Interpreter or MT output follows closely behind, maintaining fluid, real-time delivery.]}
} \\

\label{tab:examples_interpretation_scenarios}
\end{longtable}



\begin{landscape}
\end{landscape}
\end{document}